\documentclass[aoas,preprint]{imsart}
\pdfoutput=1
\RequirePackage[OT1]{fontenc}
\RequirePackage{amsthm,amsmath}
\RequirePackage[numbers]{natbib}
\usepackage{hyperref}
\hypersetup{colorlinks,citecolor=blue,urlcolor=blue}
\usepackage[utf8]{inputenc}
\usepackage{algorithm,algpseudocode,listings}
\usepackage{subcaption}

\usepackage{tikz, graphics}
\usetikzlibrary{shapes,shadows,calc}
\usepgflibrary{arrows}

\begin{document}

\begin{frontmatter}

\title{Point process modeling of drug overdoses with heterogeneous and missing data}
\runtitle{Point process modeling of drug overdoses}

\begin{aug}
\author{\fnms{Xueying} \snm{Liu}\thanksref{m1}\ead[label=e1]{xl17@iu.edu}},
\author{\fnms{Jeremy} \snm{Carter} \thanksref{m1}\ead[label=e2]{carterjg@iupui.edu}},
\author{\fnms{Brad} \snm{Ray} \thanksref{m1} \ead[label=e3]{bradray@iupui.edu}},
\author{\fnms{George} \snm{Mohler}\thanksref{m1} \ead[label=e4]{gmohler@iupui.edu}}
\runauthor{X.Liu et al.}

\affiliation{Indiana University - Purdue University Indianapolis\thanksmark{m1}} 

\address{Address of Xueying Liu and Dr. George Mohler\\
Department of Computer and Information Science\\
Indiana University - Purdue University Indianapolis \\
\printead{e1}\\
\phantom{E-mail:\ }\printead*{e4}}

\address{Address of Dr. Jeremy Carter and Dr. Brad Ray\\
O'Neill School of Public and Environmental Affairs \\
Indiana University - Purdue University Indianapolis \\
\printead{e2}\\
\phantom{E-mail:\ } 
\printead*{e3}}
\end{aug}

%%%%%%%%%%%%%%%%%%%%%%%%%%%%%%%%%%%%%%%%%%%%%%%%%%%%%%%%%%%%%%%%%%%%%%%%%%%%%%%%
\begin{abstract}
Opioid overdose rates have increased in the United States over the past decade and reflect a major public health crisis.  Modeling and prediction of drug and opioid hotspots, where a high percentage of events fall in a small percentage of space-time, could help better focus limited social and health services.  In this work we present a spatial-temporal point process model for drug overdose clustering.  The data input into the model comes from two heterogeneous sources: 1) high volume emergency medical calls for service (EMS) records containing location and time, but no information on the type of non-fatal overdose and 2) fatal overdose toxicology reports from the coroner containing location and high-dimensional information from the toxicology screen on the drugs present at the time of death.  We first use non-negative matrix factorization to cluster toxicology reports into drug overdose categories and we then develop an EM algorithm for integrating the two heterogeneous data sets, where the mark corresponding to overdose category is inferred for the EMS data and the high volume EMS data is used to more accurately predict drug overdose death hotspots.  We apply the algorithm to drug overdose data from Indianapolis, showing that the point process defined on the integrated data out-performs point processes that use only coroner data (AUC improvement .81 to .85).  We also investigate the extent to which overdoses are contagious, as a function of the type of overdose, while controlling for exogenous fluctuations in the background rate that might also contribute to clustering.  We find that drug and opioid overdose deaths exhibit significant excitation, with branching ratio ranging from .72 to .98.
\end{abstract}

\begin{keyword}
\kwd{point process}
\kwd{expectation maximization algorithm} 
\kwd{semi-supervised learning}
\kwd{non-negative matrix factorization}
\kwd{opioid overdose}
\end{keyword}

\end{frontmatter}

\section{Introduction}
Over 500,000 drug overdose deaths have occurred in the United States since 2000 and over 70,000 of these deaths occurred in 2017 \cite{seth2018overdose}. Opioids are a leading cause in these deaths and these trends are characterized by three distinct time periods \cite{ciccarone2017fentanyl}.  In the 1990s overdose deaths were driven by prescription opioid-related deaths \cite{cicero2014changing}, whereas reduced availability of prescriptions led to an increase of heroin-related deaths beginning in the 2010s \cite{cicero2014changing,rudd2014increases,strickler2019effects}. Illicit fentanyl, a synthetic opioid 50 to 100 times more potent than morphine \cite{gladden2016fentanyl}, has become a major cause of opioid-related deaths since around 2013.  It is estimated that in 2016 around half of opioid-related deaths contained fentanyl \cite{jones2018changes}, and fentanyl mixed into heroin and cocaine is likely contributing to many of these overdose deaths \cite{kandel2017increases,mccall2017recent}.

Criminology and public health disciplines have leveraged spatio-temporal event modeling in attempts to predict social harm for effective interventions \cite{mohler2018improving,tsui2011recent,yu2007interactive}. Fifty percent of crime has been shown to concentrate within just 5 percent of an urban geography \cite{weisburd2015law}.  Geographic concentrations of drug-related emergency medical calls for service \cite{hibdon2014you}, drug activity \cite{hibdon2017concentration}, and opioid overdose deaths mirror those of crime \cite{carter2018spatial}.  In particular, over half of opioid overdose deaths in Indianapolis occur in less than 5\% of the city \cite{carter2018spatial}.

Patterns of repeat and near-repeat crime in space and time further suggest that not only does crime concentrate in place but that such events are an artifact of a contagion effect resulting from an initiating criminal event \cite{townsley2003infectious}.  Similar observations have also explained the diffusion of homicide events \cite{zeoli2014homicide}. Experiments of predictive policing models using spatio-temporal Hawkes and self-exciting point processes demonstrates that such empirical realities can be harnessed to direct police resources to reduce crime \cite{mohler2015randomized}. Thus, the inter-dependence and chronological occurrence of event types in crime and public health lend promise to how to best predict other social harm events, such as opioid overdoses. 

In this work we consider the modeling of two datasets of space-time drug and opioid overdose events in Indianapolis. The first dataset consists of emergency medical calls for service (EMS) events.  These events are non-fatal overdoses and include a date, time and location, but no information on the cause of the overdose.  The second dataset consists of overdose death events (including location) and are accompanied by a toxicology report that screens for substances present or absent in the overdose event.  We develop a marked point process model for the heterogeneous dataset that uses non-negative matrix factorization to reduce the dimension of the toxicology reports to several categories.  We then use an Expectation-Maximization algorithm to jointly estimate model parameters of a Hawkes process and simultaneously infer the missing overdose category for the nonfatal overdose EMS data.  

We show that the point process defined on the integrated, heterogeneous data out-performs point processes that use only homogeneous coroner data.  We also investigate the extent to which overdoses are contagious, as a function of the type of overdose, while controlling for exogenous fluctuations in the background rate that might also contribute to clustering.  We find that opioid overdose deaths exhibit significant excitation, with branching ratio ranging from .72 to .98.

The outline of the paper is as follows.  In Section II, we provide an overview of our modeling framework.  In Section III, we run several experiments on synthetic data to validate the model and also on Indianapolis drug overdose data to demonstrate model accuracy on the application.  We discuss several policy implications and directions for future research in Section IV.

\section{Methods}

\subsection{Self-exciting point processes}

In this work we consider a self-exciting point process of the form,
~\cite{Mohler2011}:
\begin{equation}
\lambda(x,y,t) = \mu_0\nu(t)u(x,y) + \sum_{i:t_i<t} g(x-x_i,y-y_i,t-t_i),
\label{general}
\end{equation}
where $g(x,y,t)$ is a triggering kernel modeling the extent to which risk following an event increases and spreads in space and time.  The background Poisson process modeling spontaneous events is assumed separable in space and time, where $u(x,y)$ models spatial variation in the background rate and 
$\nu(t)$ may reflect temporal variation arising from time of day, weather, seasonality, etc.  The point process may be viewed as a branching process (or superposition of Poisson processes), where the background Poisson process with intensity $\mu_0\nu(t) u(x,y)$ yields the first generation and then each event $(x_i,y_i,t_i)$ triggers a new generation according to the Poisson process $g(x-x_i,y-y_i,t-t_i)$.

We allow for self-excitation in the model to capture spatio-temporal clustering of overdoses present in the data.  For example, a particular supply of heroin may contain an unusually high amount of fentanyl, leading to a cluster of overdoses in a neighborhood where the drug is sold and within a short time period.

Model \ref{general} can be estimated via an Expectation-Maximization algorithm \cite{Veen2008,mohler2011self}, leveraging the branching process representation of the model. Let $L$ be a matrix where $l_{ij}=1$ if event $i$ is triggered by event $j$ in the branching process and $l_{ii}=1$ if event $i$ is a spontaneous event from the background process.  Then the complete data log-likelihood is given by,
\begin{eqnarray}
\sum_i l_{ii}\log(\mu_0\nu(t_i)u(x_i,y_i))-\int \mu_0\nu(t)u(x,y,)dxdydt &&\\
+\sum_{ij}l_{ij}\log(g(x_i-x_j,y_i-y_j,t_i-t_j))\\-\sum_{j}\int g(x-x_j,y-y_j,t-t_j)dxdydt.    
\end{eqnarray}
Thus estimation decouples into two density estimation problems, one for the background intensity and one for the triggering kernel.  Because the complete data is not observed, we introduce a matrix
$P$ with entries $p_{ij}$ representing the probability that event $i$ is triggered by event $j$. 

Given an initial guess $P_0$ of matrix $P$, 
a non-parametric density estimation procedure can be used to estimate $u$ and $v$ from $\left\{t_k, x_k, y_k, p_{kk}\right\}_{k=1}^N$, 
providing estimates $u_0, v_0$ in the maximization step of the algorithm.

More specifically, we estimate $u$ and $v$ using leave-one-out kernel density estimation,
	\begin{equation}
	\begin{aligned}
		& v(t_i) =\frac{1}{N_b} \sum_{i \ne j}  \frac{p_{jj}}{2\pi {b_1}^2} \text{exp}\left\{- \frac{(t_i-t_j)^2 }{2 {b_1}^2}\right\}\\
		& u(x_i, y_i) = \frac{1}{N_b}\sum_{i \ne j}  \frac{p_{jj}}{2\pi {b_2}^2} \text{exp}\left\{- \frac{(x_i-x_j)^2 +(y_i-y_j)^2}{2 {b_2}^2}\right\},
		\label{kde}
	\end{aligned}
	\end{equation}
where $N_b=\sum_i p_{ii}$ is the estimated number of background events and $b_1$, $b_2$ are the kernel bandwidths that can be estimated via cross-validation or based on nearest neighbor distances.  Because $u$ and $v$ are chosen to integrate to 1, we then have the ML estimate $\hat{\mu}_0=N_b$

 \begin{figure}[h]
 \centering
\includegraphics[width=0.6\textwidth]{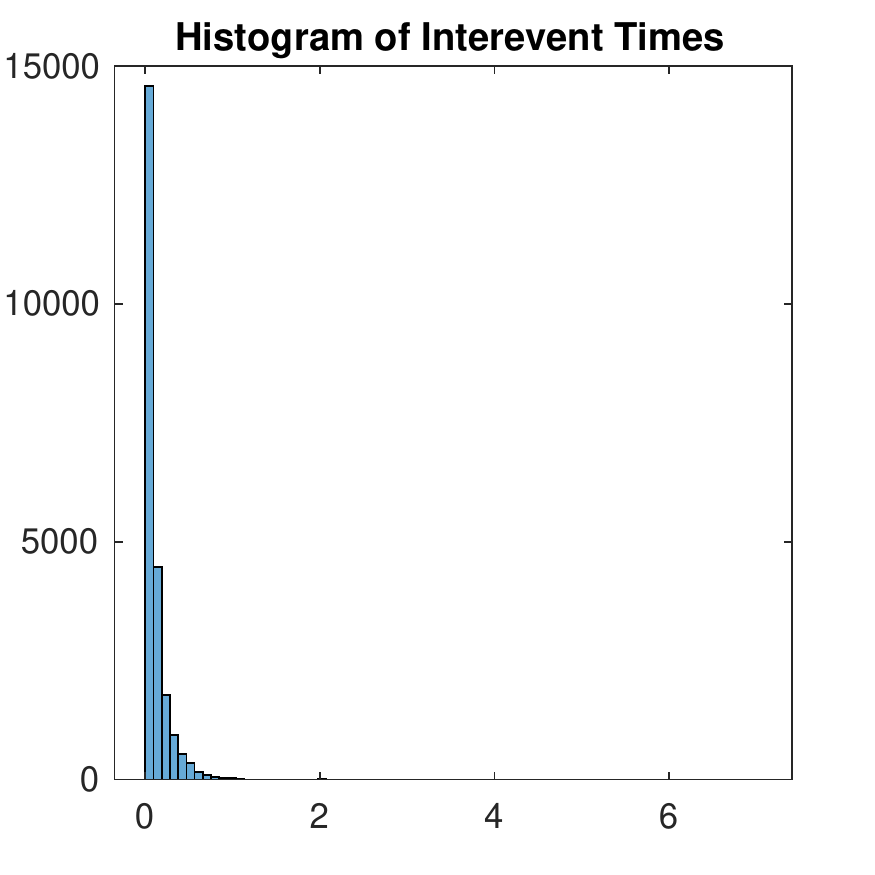}
\caption{Histogram of inter-event times of real data, suggests that time triggering function is exponential.}
\label{fig:interevent}
\end{figure}

We assume the triggering kernel is given by a separable function that is exponential in time (Figure \ref{fig:interevent}) with parameter $\omega$ and Gaussian in space with parameter $\sigma$ \cite{mohler2014marked},
\begin{eqnarray}
    g(x,y,t)= & K_{0} \left(w\cdot \text{exp}\left\{-w t\right\}\right) \cdot \\
& \frac{1}{{2\pi {\sigma}^2}}\cdot \text{exp} \left\{-\frac{1}{2{\sigma}^2} ( x^2+y^2)\right\}.
\end{eqnarray}
We then obtain an estimate for the parameters using weighted sample averages from the data $\left\{t_i - t_j, x_i - x_j, y_i - y_j, p_{ij}\right\}_{t_i>t_j}$,
\begin{equation}
		\begin{aligned}
	&	\hat{K}_{0} =  \sum_{t_i > t_j} p_{ij}\bigg/\sum_{i,j}p_{ij}, \\
	&	\hat{w} =  \sum_{t_i > t_j} p_{ij}\bigg/\sum_{t_i > t_j} p_{ij} \cdot (t_i-t_j),\\
	&	\hat{\sigma} = \sqrt{\sum_{t_i>t_j} p_{ij} \cdot \left[ (x_i-x_j)^2 + (y_i-y_j)^2\right]\bigg/2\cdot \sum_{t_i > t_j} p_{ij}}
		\end{aligned}
			\label{kde2}
\end{equation}

In the estimation step, we estimate the probability that event $i$ is a background event via the formula,
\begin{equation}
p_{ii} = \frac{\mu_0 u(x_i,y_i)v(t_i)}{\lambda(x_i,y_i,t_i)},
\label{pii}
\end{equation}
and the probability that event $i$ is triggered by event $j$ as,
\begin{equation}
p_{ij} = \frac{g(x_i-x_j, y_i-y_j, t_i-t_j)}{\lambda(x_i, y_i, t_i)},
\label{pij}
\end{equation}
\cite{Zhuang2002}.
We then iterate for $n=1,...,N_{em}$ between the expectation and maximization steps until a convergence criteria is met: 
\begin{enumerate}
    \item Estimate $u_n, v_n$, and $g_n$ using (\ref{kde}) and (\ref{kde2}).
    \item Update $P_n$ from  $u_n, v_n$, and $g_n$ using (\ref{pii}) and (\ref{pij}).
\end{enumerate}

\subsection{Modeling with heterogeneous event data}
In this work we assume that we are given two datasets $A$ and $B$, though our modeling framework extends more generally to three or more.  Event dataset $A$ contains low dimensional, unmarked space-time events, whereas dataset $B$ contains space-time events with high-dimensional marks.  In our application, drug overdoses that do not result in death comprise dataset $A$, whereas those overdoses that do result in death are accompanied by a high-dimensional mark, namely the toxicology screen conducted by the coroner.  Event dataset $B$ therefore contains a much smaller number of events compared to $A$.

Next we use non-negative matrix factorization (NMF) \cite{lee2001algorithms} to reduce the dimension of the high-dimensional mark of dataet $B$ into an indicator for $K$ groups. Each toxicology report consists of an indicator (presence or absence) for each one of 133 drugs the test screens.  These reports then are input into a overdose-drug matrix analogous to a document-term matrix in text analysis using NMF.  We then use NMF to factor overdose-drug matrices into the product of two non-negative matrices, one of them representing the relationship between drugs and topic clusters and the other one representing the relationship between topic clusters and specific overdose events in the latent topic space.  The second matrix yields the cluster membership of each event (the cluster is the argmax of the column corresponding to each event).

\subsection{Estimation of a marked point process with missing data}

Merging dataset $A$ and $B$, we now have marked event data $(x_i,y_i,t_i,k_i)$ where the mark $k_i$ is one of $k=1,...,K$ clusters and is unknown for event data coming from $A$ but is known for event data from $B$.   

Model (\ref{general}) can be extended by adding in the group labels
\begin{equation}
\lambda^k(x,y,t) = \mu^k_0 u^k(x,y)v^k(t) + \sum_{\substack{i:t_i<t\\k_i=k}} g^k(x-x_i, y-y_i, t-t_i),
\label{lambdak}
\end{equation}
where $g^k$ is modelled as follows:
\begin{equation}
\begin{aligned}
g^k(x,y,t) =&  K^k_{0} \left(w^k\cdot \text{exp}\left\{-w^k t\right\}\right) \cdot \\
& \frac{1}{{2\pi {\sigma^k}^2}}\cdot \text{exp} \left\{-\frac{1}{2{\sigma^k}^2} \left[ x^2 + y^2\right]\right\},
\end{aligned}
\label{triggerk}
\end{equation}
Here we assume each cluster $k$ has its own parameters $(\omega^k,\mu_0^k,\sigma^k,K_0^k)$.

We then extend the branching structure matrix $P$ to a set of $K$ matrices, $P^k$, with initial guess $P^k_0$ and entries:
$$p^k_{ij} = \begin{cases} \frac{1}{K}, & \mbox{if } i=j \mbox{ and event } i \mbox{ from }A  \\ 1, &  \mbox{if } i= j \mbox{, event } i\mbox{ from } B \mbox{ and belongs to group } k \\ 0, & \mbox{otherwise}\end{cases}
$$
Then $P^k$ can be updated similarly for each cluster $k = 1, \cdots, K$:
\begin{equation}
p^k_{ii} = \frac{u^k(x_i,y_i)v^k(t_i)}{\lambda^k(x_i,y_i,t_i)},
\label{piik}
\end{equation}
and 
\begin{equation}
p^k_{ij} = \frac{g^k(x_i-x_j, y_i-y_j, t_i-t_j)}{\lambda^k(x_i, y_i, t_i)},
\label{pijk}
\end{equation}
where for each event $i$ from dataset $A$, we have that $\sum\limits_{k=1}^K \left(\sum\limits_{t_i \ge t_j}p^k_{ij}\right) = 1$, 
and for event $i$ from dataset $B$ we have that $p^{\Tilde{k}}_{ij} = 0$ for all events $j$ where $t_i \ge t_j$ and $\Tilde{k}$ is not the group to which event $i$ belongs.

The parameters are then estimated using $P^k$:
\begin{equation*}
	\begin{aligned}
	&	K^k_{0} =  \sum_{t_i > t_j} p^k_{ij}\bigg/\sum_{i,j}p^k_{ij}, \\
	&	w^k =  \sum_{t_i > t_j} p^k_{ij}\bigg/\sum_{t_i > t_j} p^k_{ij} \cdot (t_i-t_j),\\
	&	\sigma^k = \sqrt{\sum_{t_i>t_j} p^k_{ij} \cdot \left[ (x_i-x_j)^2 + (y_i-y_j)^2\right]\bigg/\left(2\cdot \sum_{t_i > t_j} p^k_{ij}\right)},  \\ 
	&	\mu^k_{0} = \sum p^k_{ii}.
	\end{aligned}
\end{equation*}
and the EM algorithm is iterated to convergence.
\section{Results}
\subsection{Synthetic Data}
\begin{figure}[ht]
\centering
    \begin{tikzpicture}[scale = 1]
	\begin{scope} [fill opacity = .4]
    \draw (-4,4) rectangle (0,0);
    \node at (-3.8,-0.2) {0};
    \node at (-4.2, 0.2) {0};
    \node at (-0.2, -0.2) {1};
    \node at (-4.2, 3.8) {1};
    \node at (-3,3) {bg(1)};
    \node at (-1,3) {bg(2)};
    \node at (-3,1) {bg(3)};
    \node at (-1,1) {bg(4)};
    \draw [dashed] (-4,2) -- (0,2);
    \draw [dashed] (-2,4) -- (-2,0);
    \draw [fill=blue] (-4,4) rectangle (-2,2);
    \draw [fill=yellow] (-2,2) rectangle (0,0);
    \draw [fill=green] (-4,2) rectangle (-2,0);
    \draw [fill=orange] (-2,4) rectangle (0,2);
    \end{scope}
\end{tikzpicture}
\caption{Simulation of events' location: for each background event, probabilities of falling in the purple, orange, green, and yellow regions are $bg(1)$, $bg(2)$, $bg(3)$, $bg(4)$, respectively. }
\label{simu}
\end{figure}
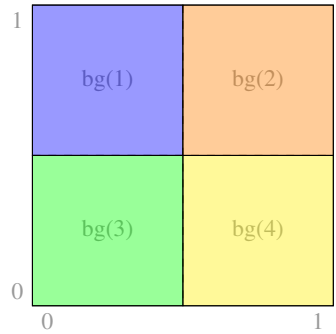

To validate our methodology, we simulate point process data where data set $B$ has $K=4$ groups with parameters given by those in Table \ref{bgsetting} and \ref{paramsetting}.  The background rate for each group is heterogeneous in space, with different rates in each quadrant in the unit square and homogeneous in time.  Figure \ref{simu} and Table \ref{bgsetting} illustrate how the background events are simulated: different background rates are assigned to each of the four different regions.  Table {\ref{paramsetting}} contains the true parameters for each group.

\begin{table}
\centering
 \begin{tabular}{l | r | r | r | r }
 \hline
 group & bg(1)  & bg(2) & bg(3) & bg(4) \\
 \hline
 1 & 0.1 & 0.2 & 0.3 & 0.4 \\
 2 & 0.4 & 0.3 & 0.2 & 0.1 \\
 3 & 0.4 & 0.4 & 0.1 & 0.1 \\
 4 & 0.1 & 0.4 & 0.1 & 0.4 \\
 \hline
 \end{tabular}
 \caption{Background rates of synthetic data.}
 \label{bgsetting}
 \end{table}

 \begin{table}
\centering
 \begin{tabular}{l | r | r | r | r }
 \hline
 group & w  & $K_0$ & $\sigma$ & $\mu$ \\
 \hline
 1 & 0.1 & 0.9 & 0.01 & 67 \\
 2 & 0.5 & 0.8 & 0.001 & 28 \\
 3 & 1 & 0.6 & 0.02 & 55 \\
 4 & 0.3 & 0.75 & 0.003 & 132 \\
 \hline
 \end{tabular}
 \caption{True parameters of synthetic data.}
 \label{paramsetting}
 \end{table}

We then simulate the missing data process by assigning $30\%$ of the data to dataset $A$ (no label) and $70\%$ to $B$. We find that the EM algorithm detailed above converges within 50 iterations.

\begin{figure*}[ht]
\centering
\includegraphics[width = .45\textwidth]{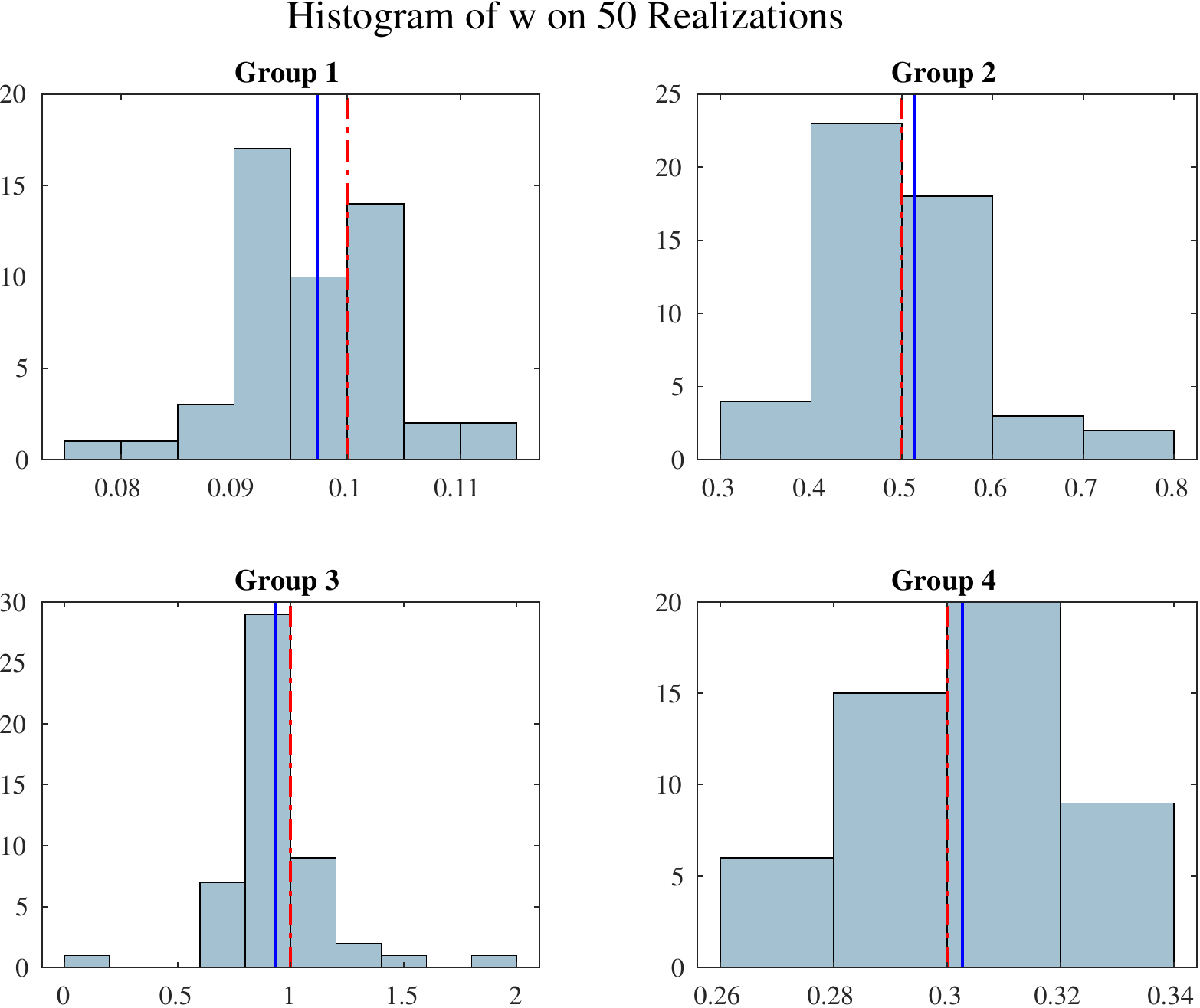}
\includegraphics[width = .45\textwidth]{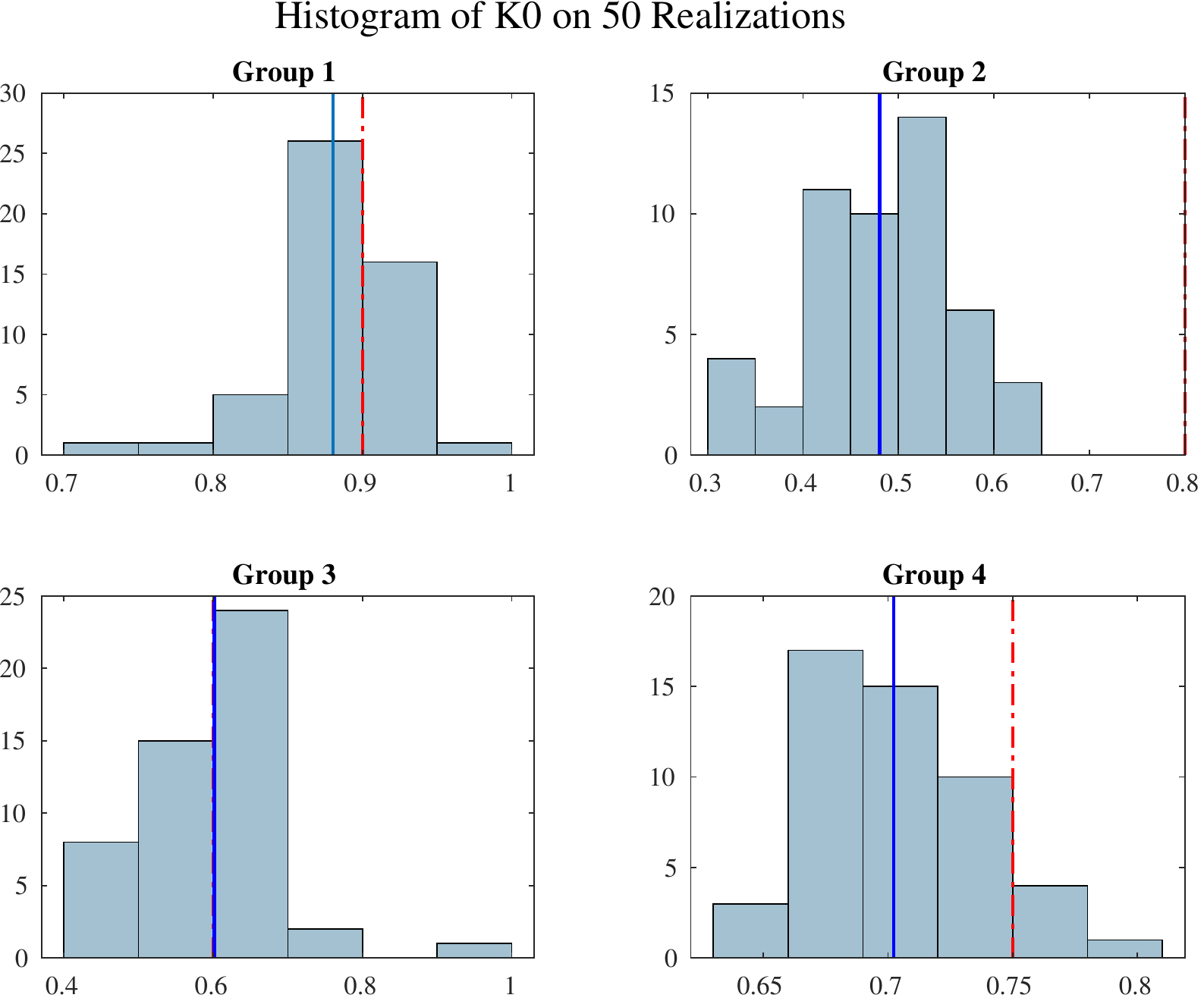}
\includegraphics[width = .45\textwidth]{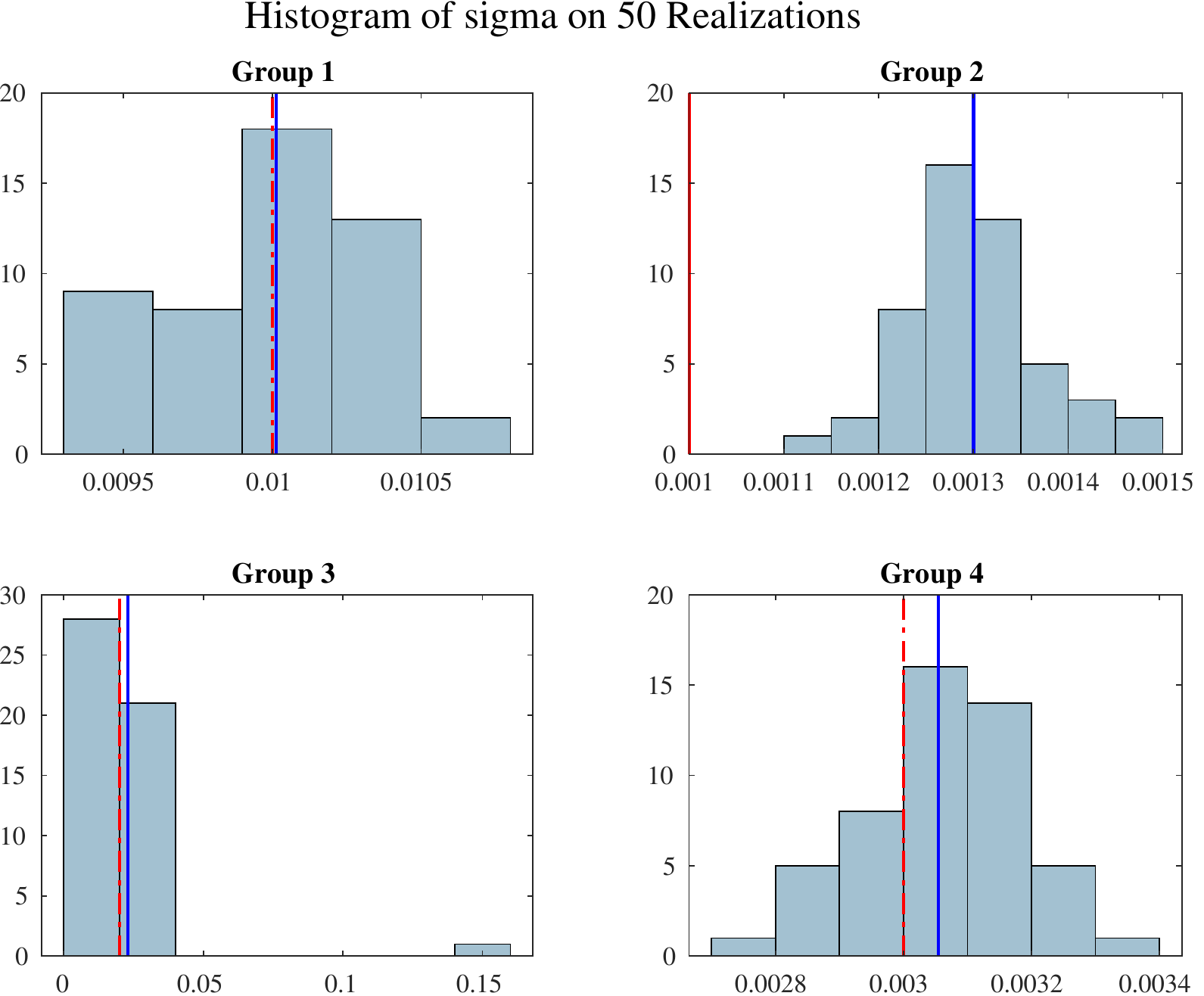}
\includegraphics[width = .45\textwidth]{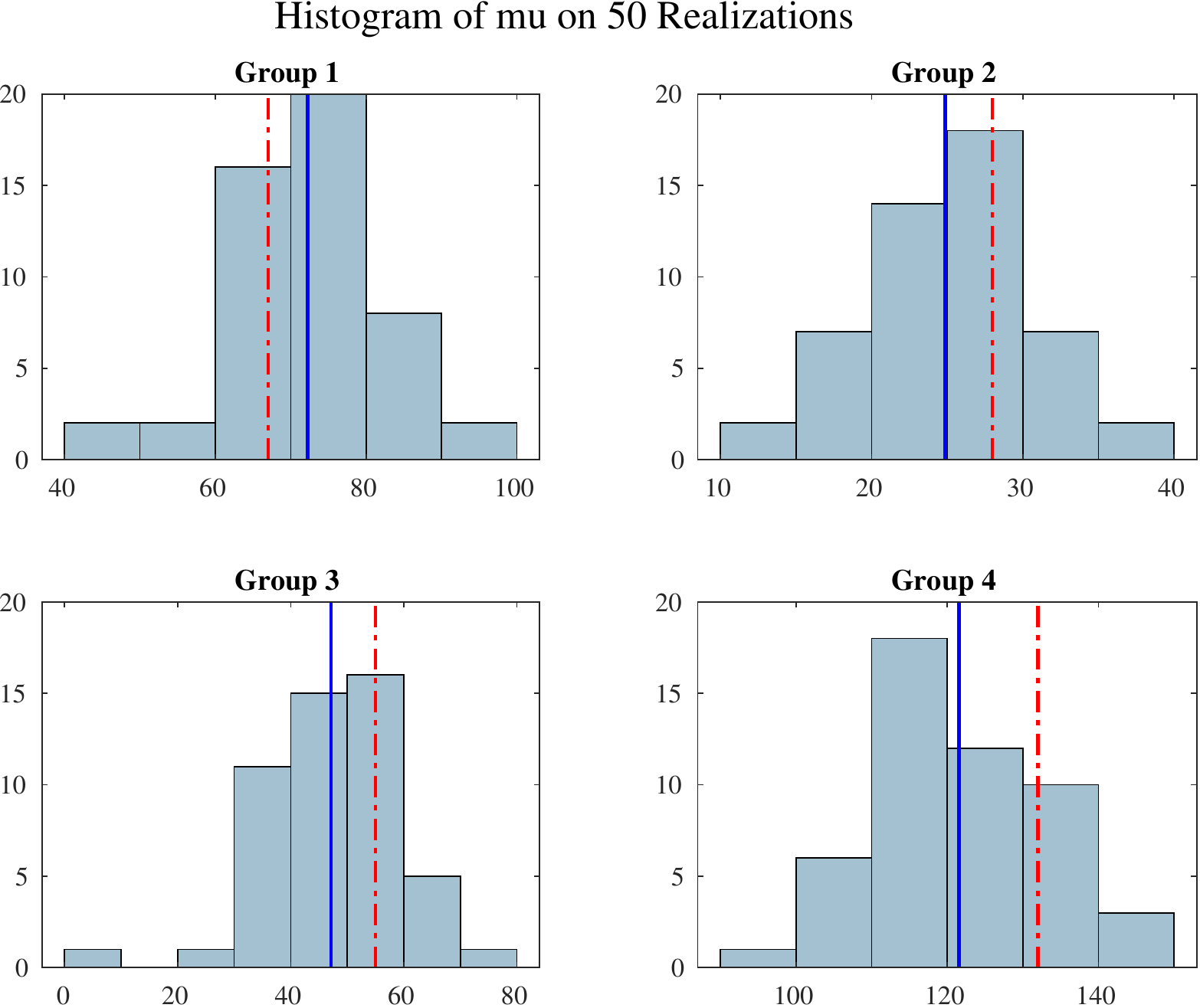}
\caption{Parameters' true value (in red dash-dot line) and average of converged values (in blue solid line).}
\label{fig:params}
\end{figure*}

We simulate 50 synthetic datasets and then estimate the true parameters, where the results are displayed in Figure \ref{fig:params}. In the figure, the histograms of $w$, $K_0$, $\sigma$ and $\mu$ correspond to the estimates from the EM algorithm, 
where the red reference lines represent the average of the 50 results and the true value of the parameters are in blue.  We find that our model is able to accurately recover both the true parameters and the event cluster membership up to the standard errors of the estimators.

In Table \ref{eventtable}, we display the estimated number of events of each group (along with their actual values) when A has 30\% of events as well as when 90\% of events are assigned to $A$ (and thus unknown).  We find in both experiments that the model is able to recover the cluster sizes accurately.
\begin{table}
\centering
 \begin{tabular}{l | r | r}
 \hline
 group & \shortstack{true \\ \#}  & \shortstack{estimated \\  \#} \\
 \hline
 group 1 & 570 & 581 \\
 group 2 & 154 & 145 \\
 group 3 &  173 & 168\\
 group 4 & 431 & 434 \\
 \hline
 \end{tabular}
 \begin{tabular}{l | r | r}
 \hline
 group &\shortstack{true \\ \#}  & \shortstack{estimated \\  \#} \\
 \hline
 group 1 & 1197& 1195 \\
 group 2 & 71 & 56 \\
 group 3 &  134 & 113\\
 group 4 & 380 & 418 \\
 \hline
 \end{tabular}
 \caption{Number of events from each group vs estimated number while dataset $A$ is $30\%$ (left) and $90\%$ (right) of all data.}
 \label{eventtable}
 \end{table}
 
\begin{figure}[h]
\centering
\includegraphics[width = 0.45\textwidth]{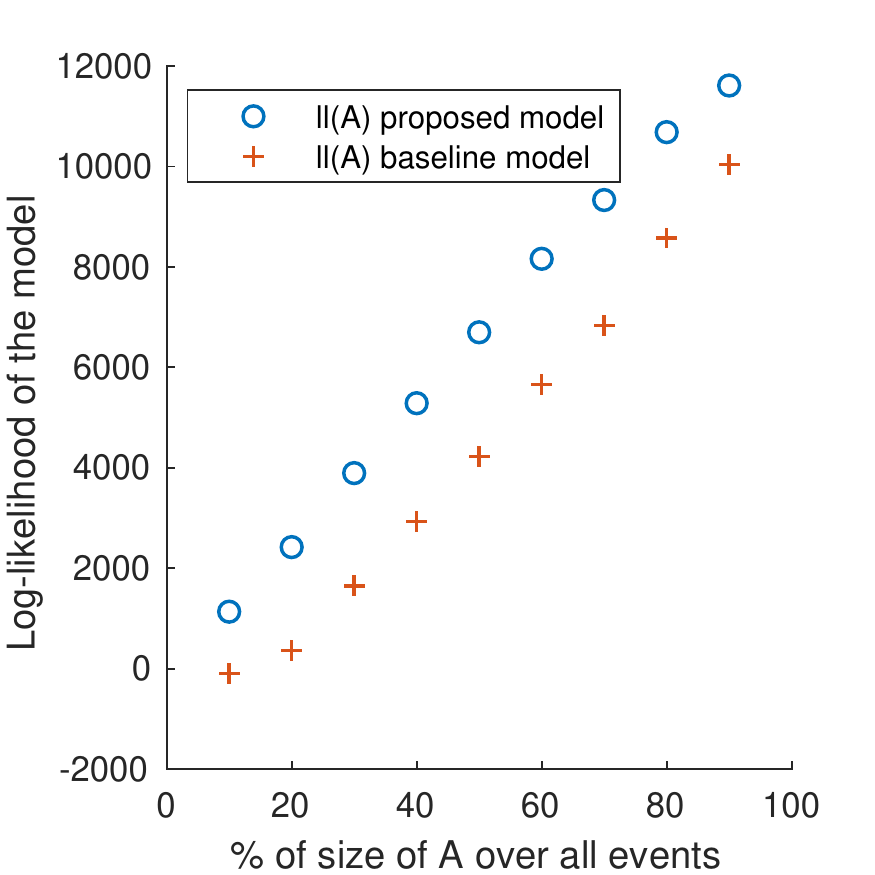}
\includegraphics[width = 0.45\textwidth]{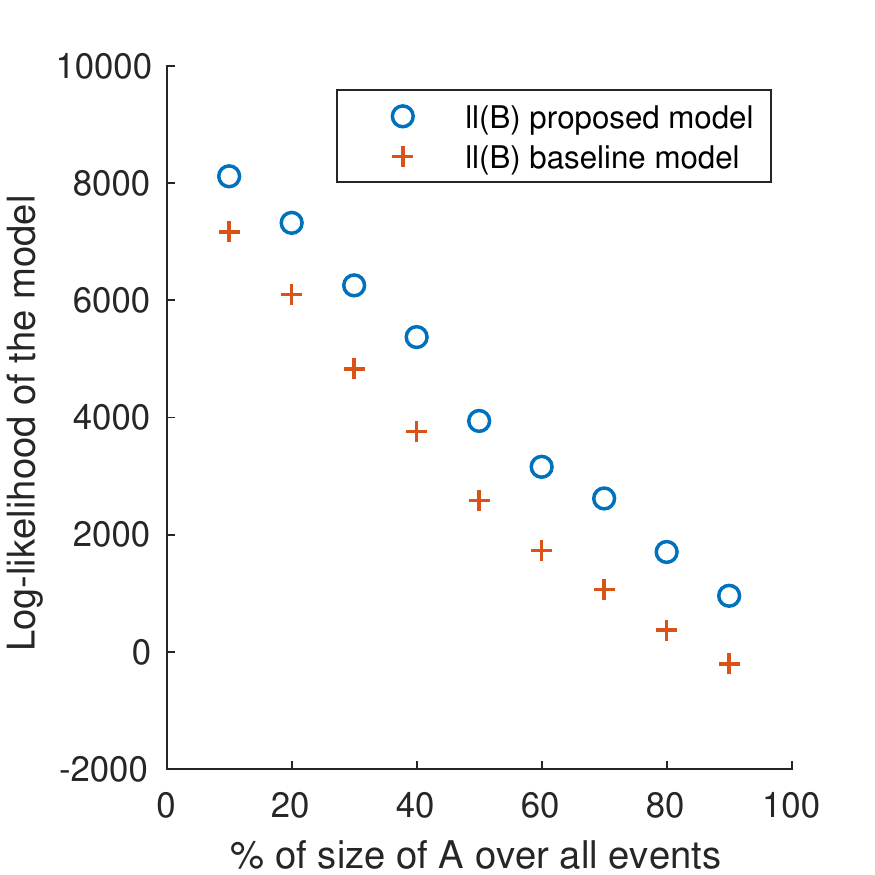}
	\caption{Log-likelihood of the model vs baseline model on individual datasets with different percentage of $A$.  Left: likelihood evaluated on dataset $A$.  Right: likelihood evaluated on dataset $B$.}
	\label{ll}
\end{figure}
 In Figure \ref{ll}, we compare baseline models estimated only on $A$ or $B$ individually against the combined model.  We also analyze the difference in performance versus the percentage of events assigned to dataset $A$. Here we find that the model estimated on both datasets always has higher likelihood than the models estimated only on one dataset.
 
 \subsection{Emergency Data and Toxicology Report}
 Next we analyze a dataset of drug overdose data from Marion County, Indiana (Indianapolis).  The data spans the time period from January 14, 2010 to December 30, 2016. 
 The fatal drug overdose dataset with toxicology reports (dataset $B$) consists of 969 events and the non-fatal, emergency medical calls for service dataset is 24 times bigger, with 22,049 unlabelled events. 
 
\begin{figure}[h]
 \centering
\includegraphics[width=0.6\textwidth]{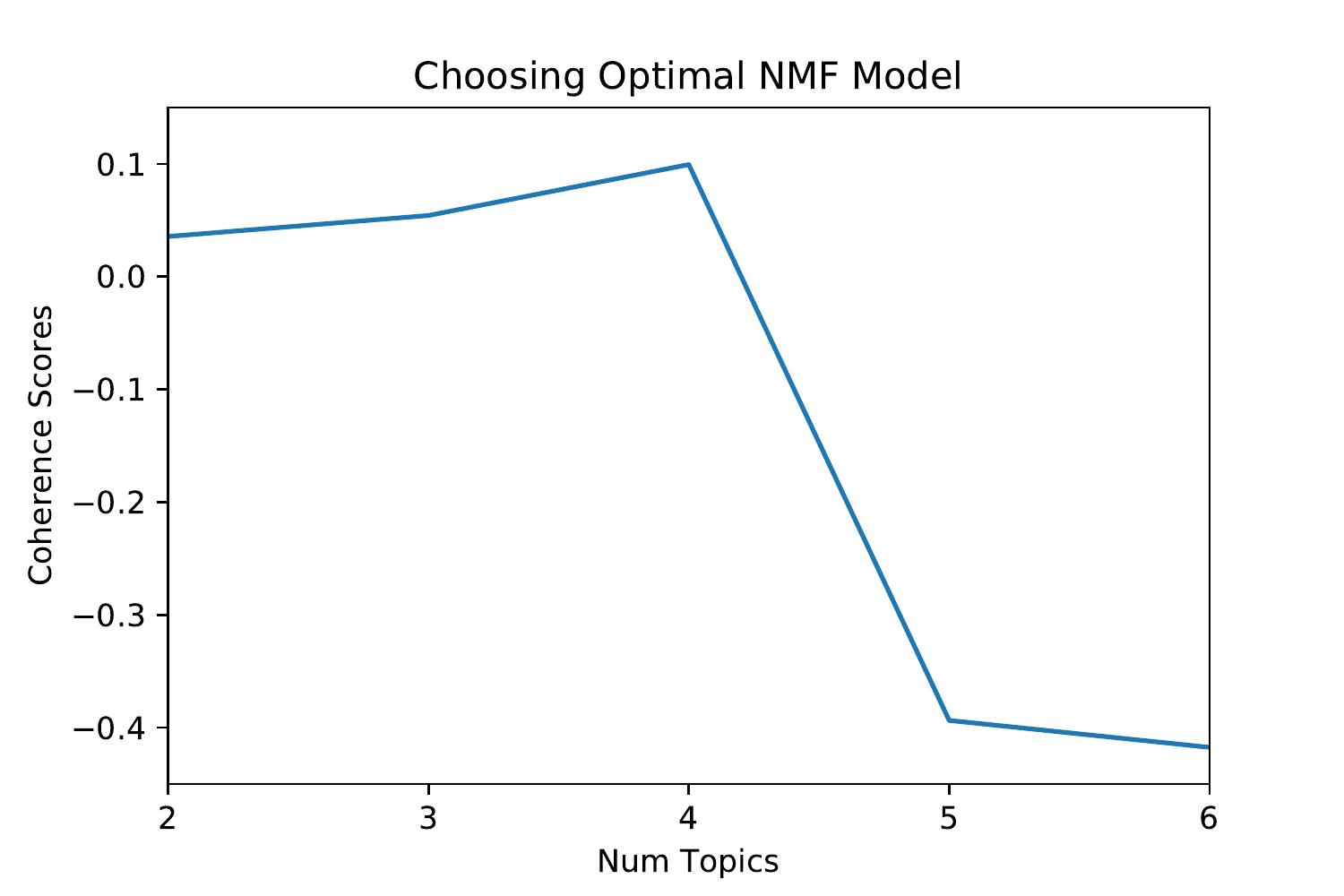}
\caption{NMF coherence scores of drug overdose clusters vs number of topic clusters $K$.}
\label{fig:NMF}
\end{figure}

We use NMF as described above to cluster the toxicology report data.  We use coherence \cite{stevens2012exploring} to select the number of clusters, which we find to be $K=4$ for our data (see Figure \ref{fig:NMF}).   In Table {\ref{24topopioid}} we show the top 24 most frequent drugs and their frequencies present in the fatal overdose dataset and  in Table {\ref{topictable}} we display the top 5 most frequent drugs found in each NMF group.  In Table {\ref{topictable}} we find that the first group consists of illicit drugs (6-MAM and heroin), whereas group 2 consists of mostly opioids that can be obtained via a prescription.  Group 3 overdoses involve alcohol, whereas group 4 is fentanyl related overdoses. 

\begin{table}
\centering
 \begin{tabular}{l | l | l | l}
 \hline
 drug & frequency & drug & frequency \\
 \hline
Hypnotic & 0.9617 & 11-Nor-9-carboxy-THC & 0.8113 \\
Lidocaine &	0.5588 & 11-Hydroxy-THC &	0.4856 \\
Phenobarbital &	0.4762 & Gastrointestinal	& 0.3841 \\
Eszopiclone & 0.3841  & THC-Aggregate &	0.3580\\
Promethazine &	0.3566 & Alcohol	& 0.2451 \\
Ethanol	& 0.2451 & Opioids & 0.2263 \\
Illicit &	0.2189 & Norfentanyl &	0.1773 \\
Amphetamine & 0.1760 & Acetylfentanyl & 0.1605 \\
Fentanyl &	0.1571 & Acetyl	&0.1343 \\
Methamphetamine	&0.1162 & Morphine	&0.1162 \\
Delta-9-THC	&0.0907 &6-MAM	&0.0604 \\
Diazapam	&0.0537 & THC	&0.0524 \\
\hline
\end{tabular}
\caption{24 most frequently present drugs.}
\label{24topopioid}
\end{table}

\begin{figure}[h]
 \centering
    \begin{subfigure}[b]{0.7\textwidth}
	\includegraphics[width=\textwidth]{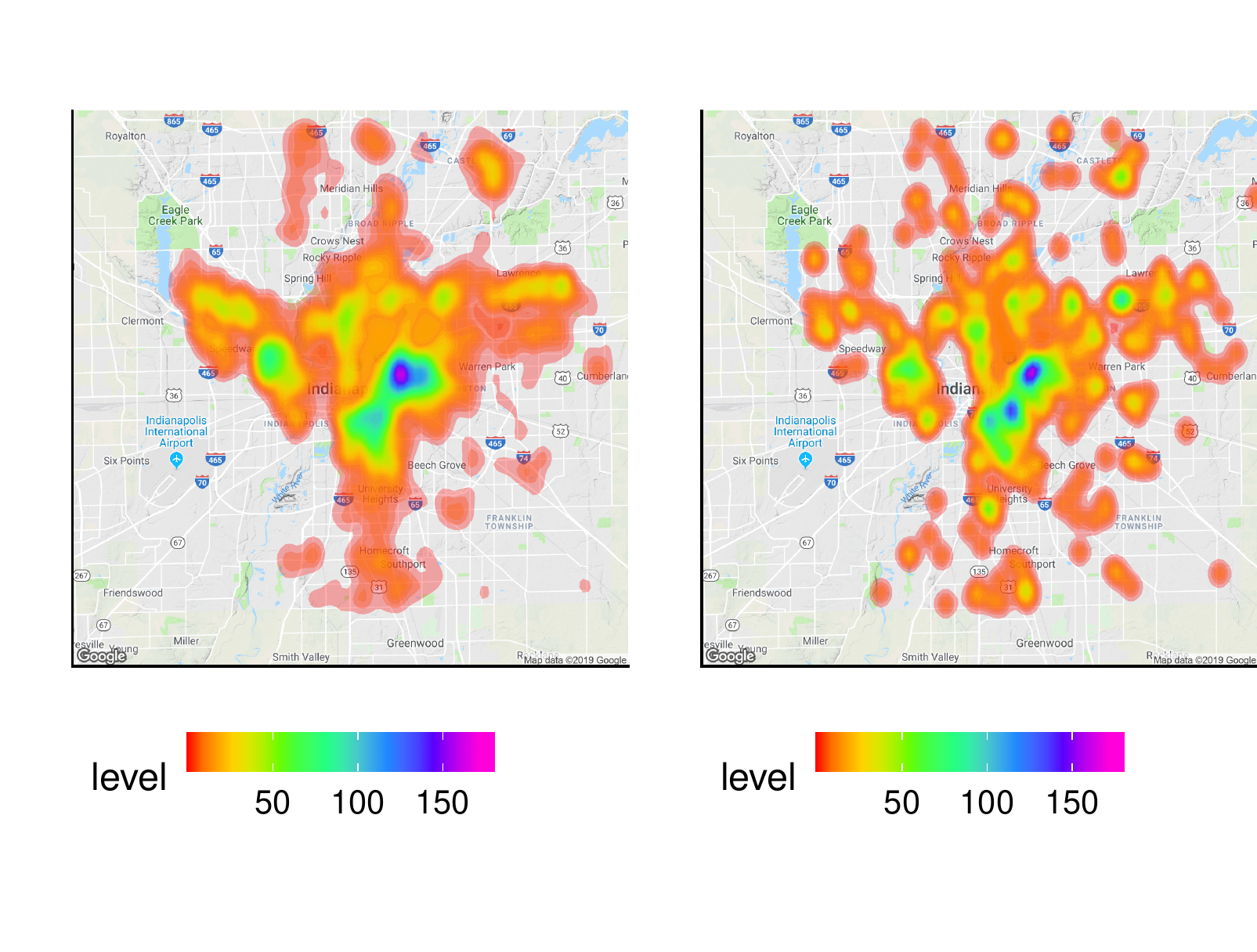}
	\label{fig:hm1}
	\end{subfigure}
	
	\vspace{-10mm}
	\begin{subfigure}[b]{0.7\textwidth}
	\includegraphics[width=\textwidth]{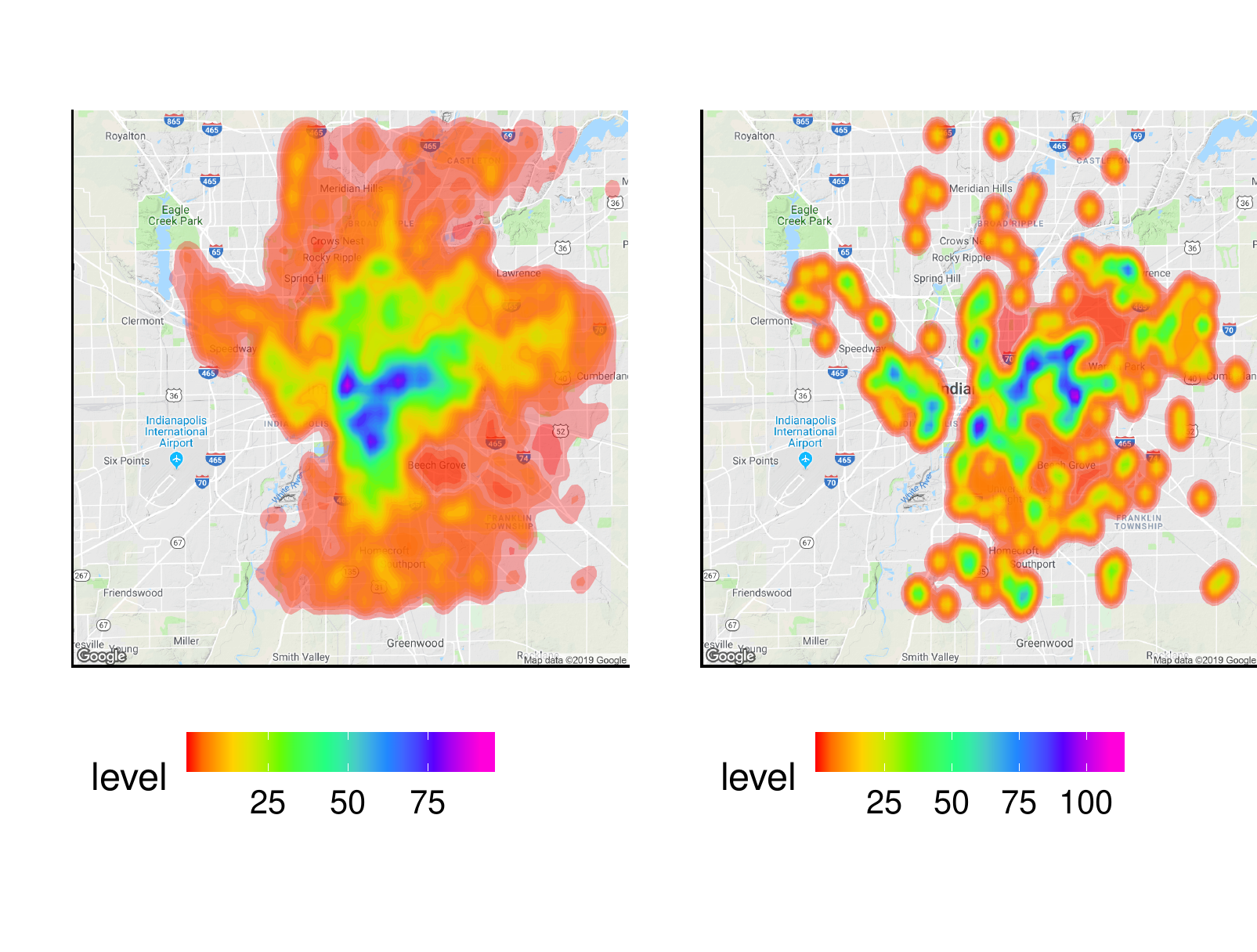}
	\label{fig:hm2}
	\end{subfigure}
	
	\vspace{-10mm}
    \begin{subfigure}[b]{0.7\textwidth}
	\includegraphics[width=\textwidth]{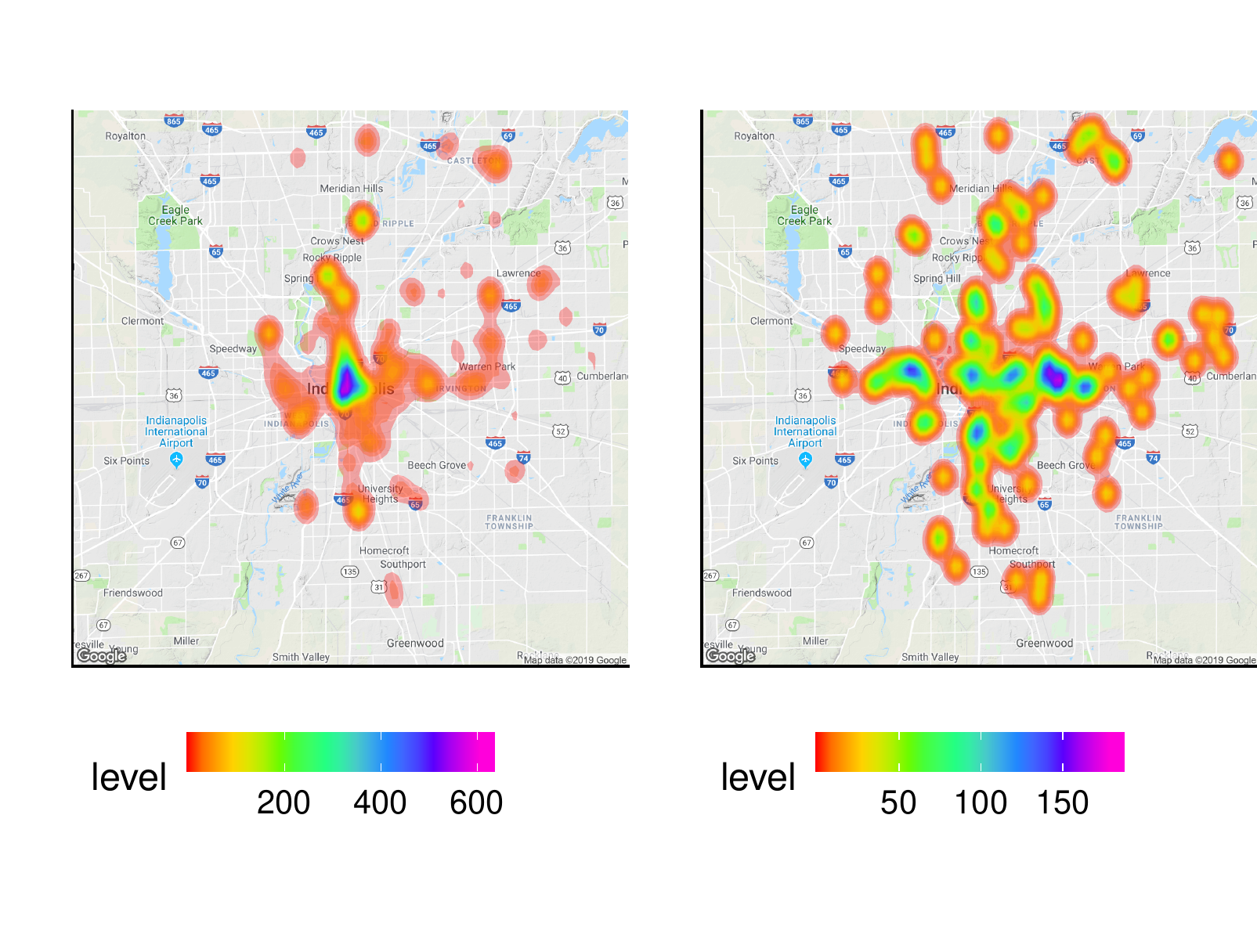}
	\label{fig:hm3}
	\end{subfigure}
	
	\vspace{-10mm}
	\begin{subfigure}[b]{0.7\textwidth}
	\includegraphics[width=\textwidth]{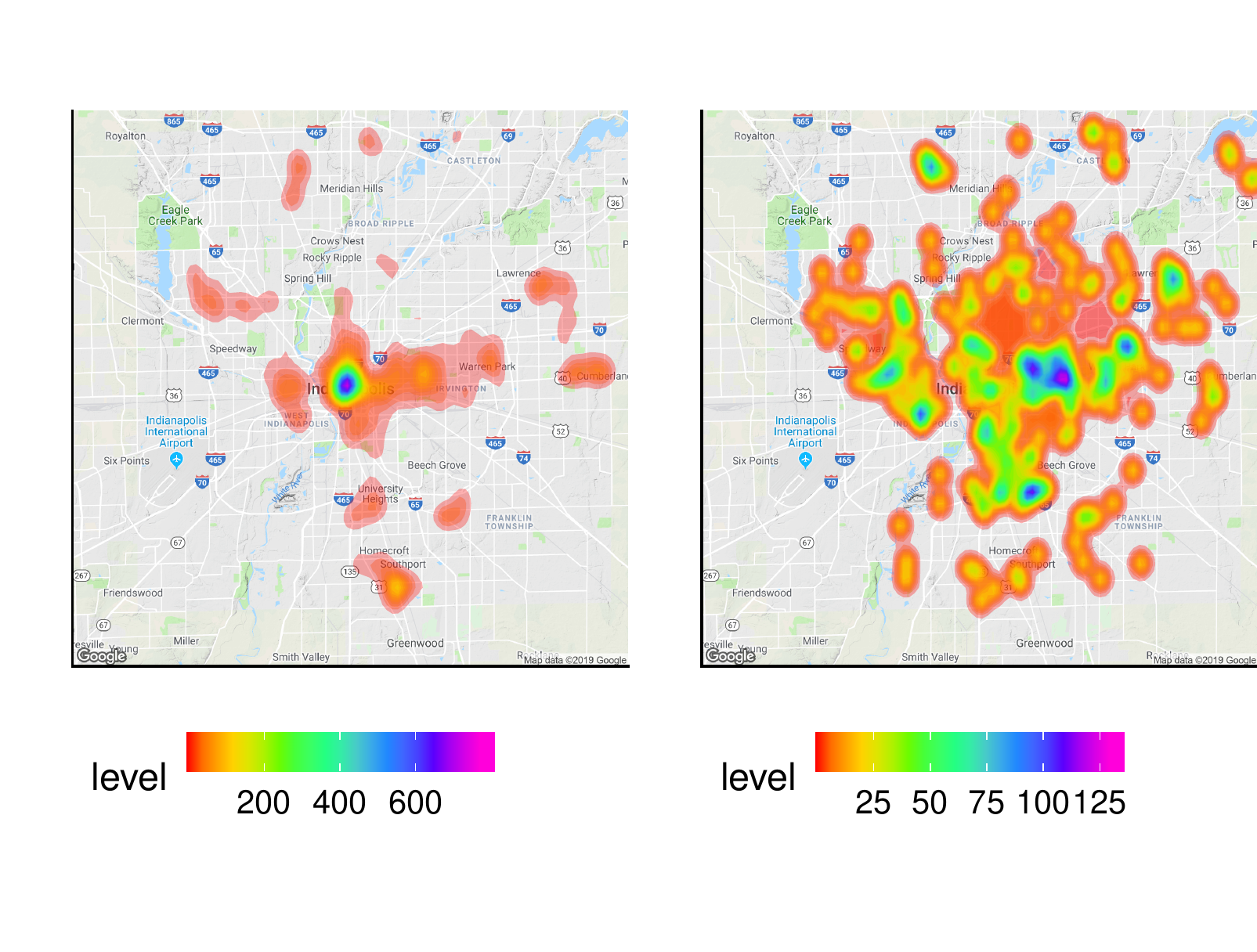}
	\label{fig:hm4}
	\end{subfigure}
	\caption{Heatmaps of non-fatal overdose events (left) and fatal overdose events (right).  Top to bottom: groups 1-4. }
	\label{fig:hms}
\end{figure}

\begin{figure}[h]
\centering
\begin{subfigure}[b]{0.4\textwidth}
	\includegraphics[width=\textwidth]{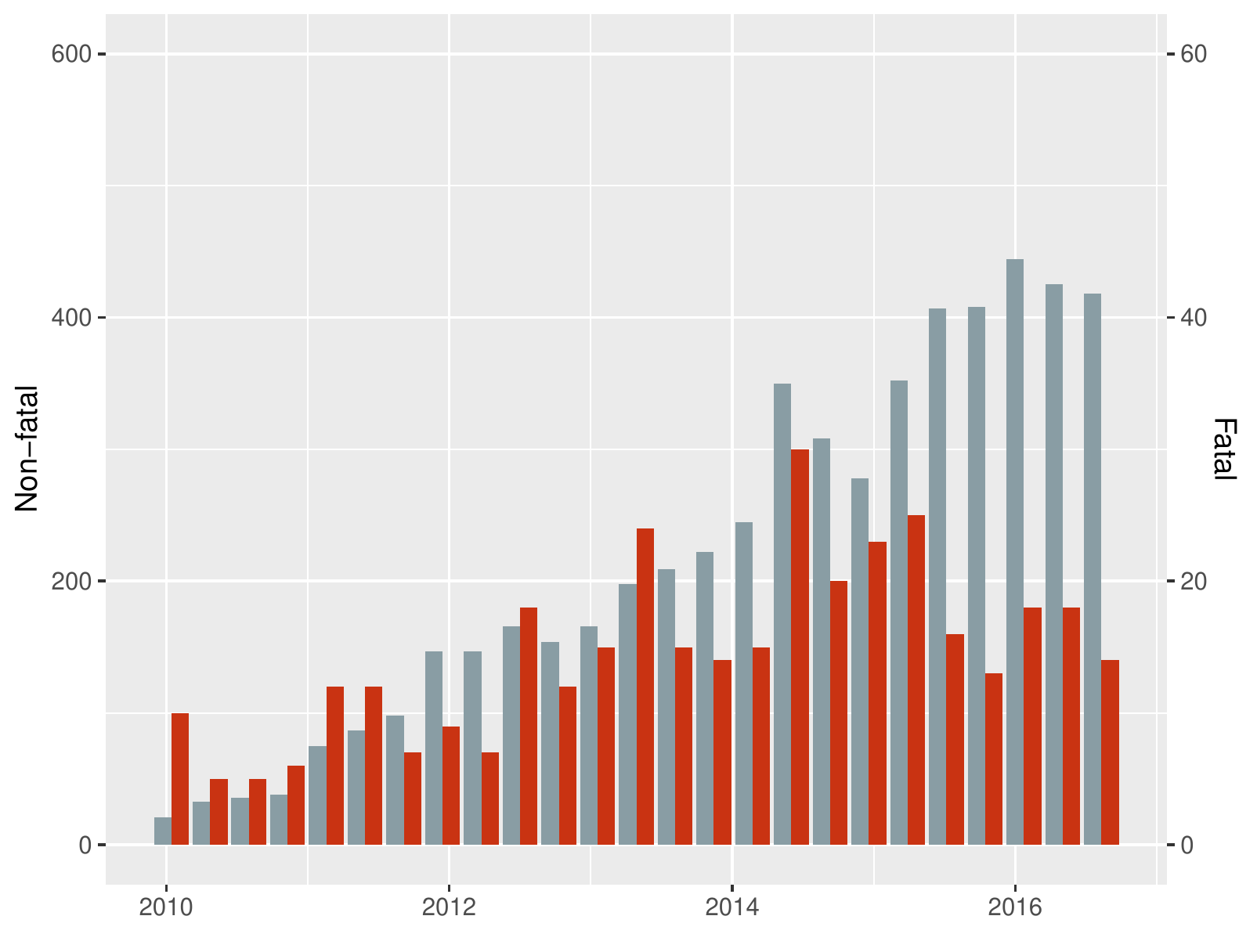}
	\label{fig:hist1}
	\end{subfigure}
	\begin{subfigure}[b]{0.4\textwidth}
	\includegraphics[width=\textwidth]{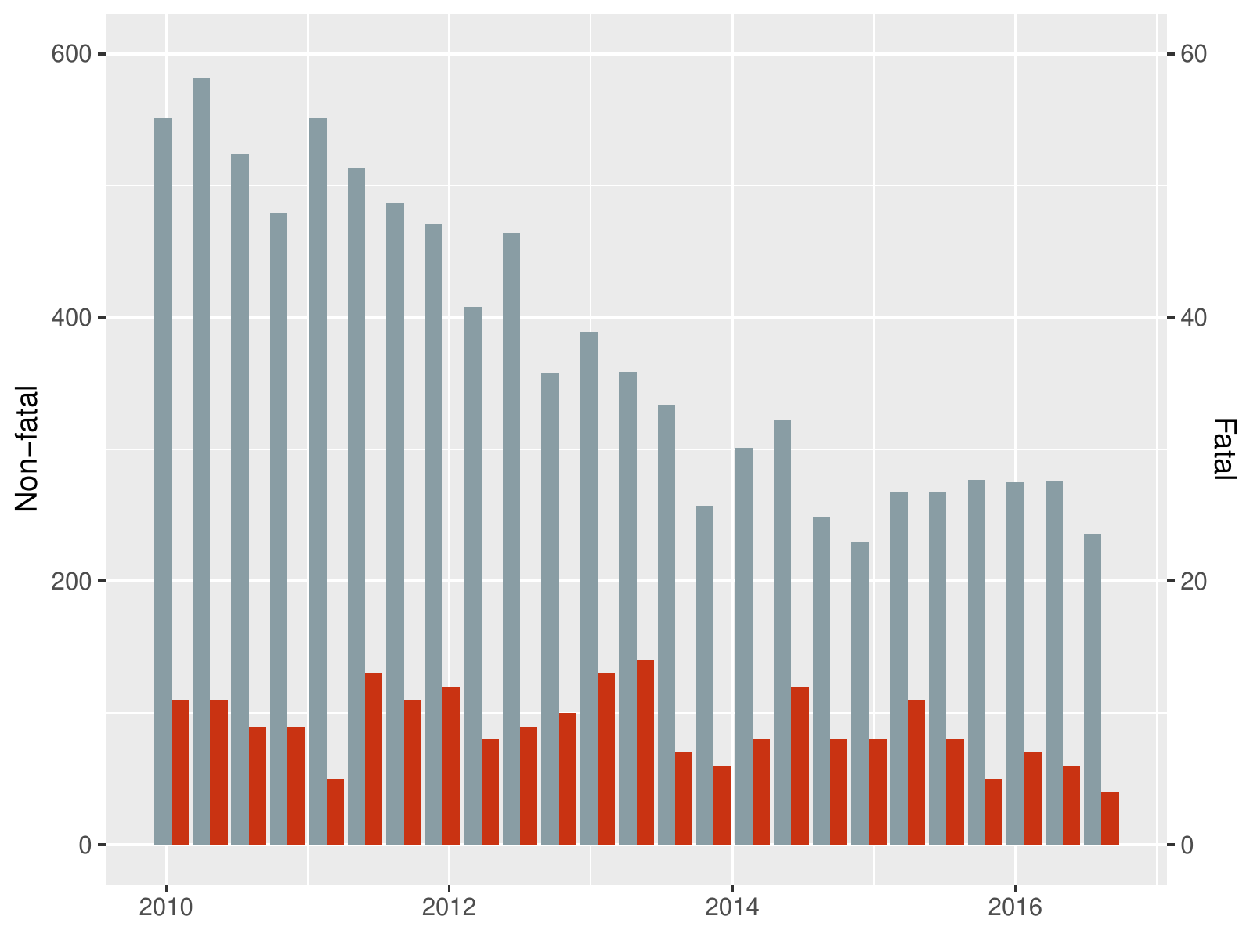}
	\label{fig:hist2}
	\end{subfigure}
    \begin{subfigure}[b]{0.4\textwidth}
	\includegraphics[width=\textwidth]{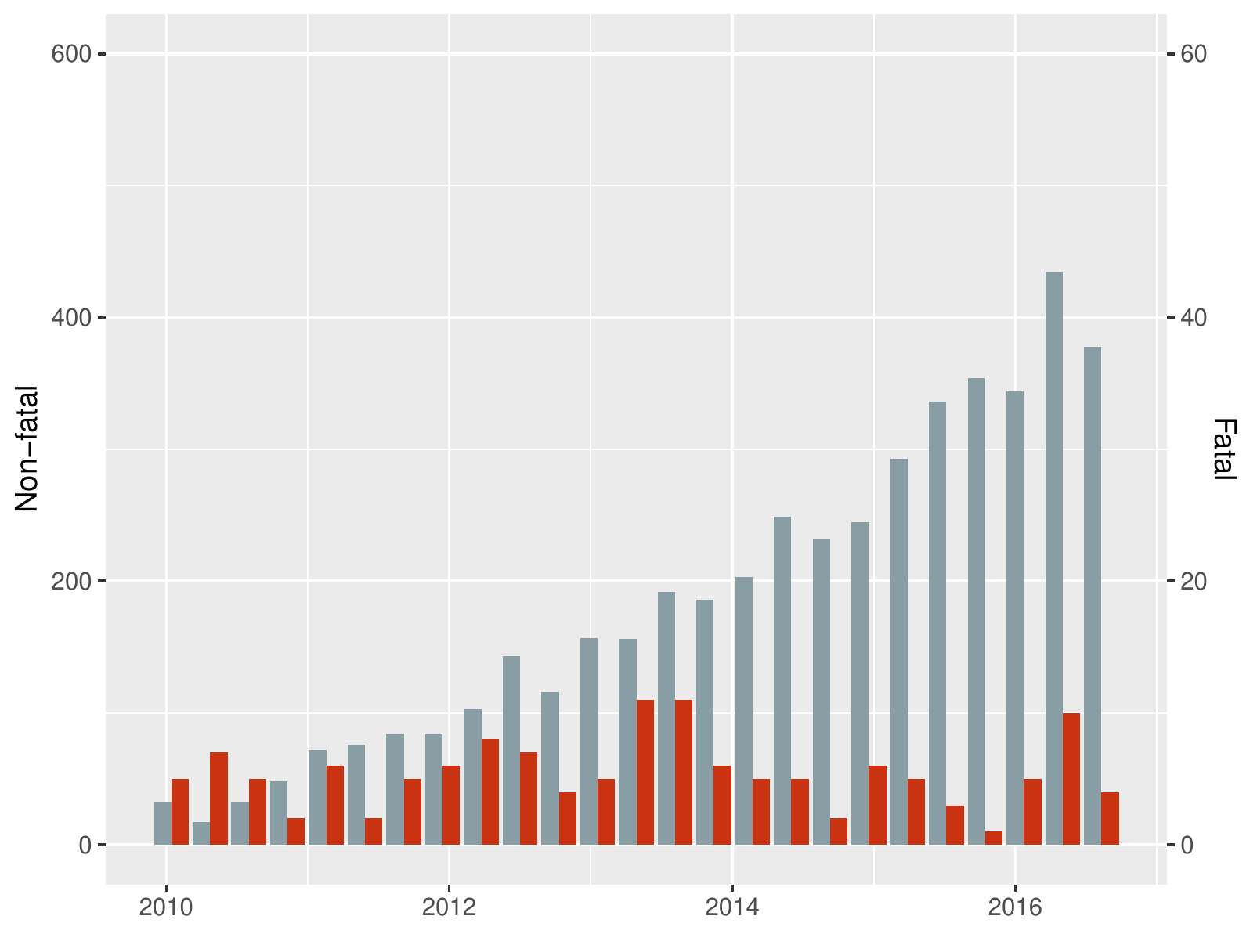}
	\label{fig:hist3}
	\end{subfigure}
	\begin{subfigure}[b]{0.4\textwidth}
	\includegraphics[width=\textwidth]{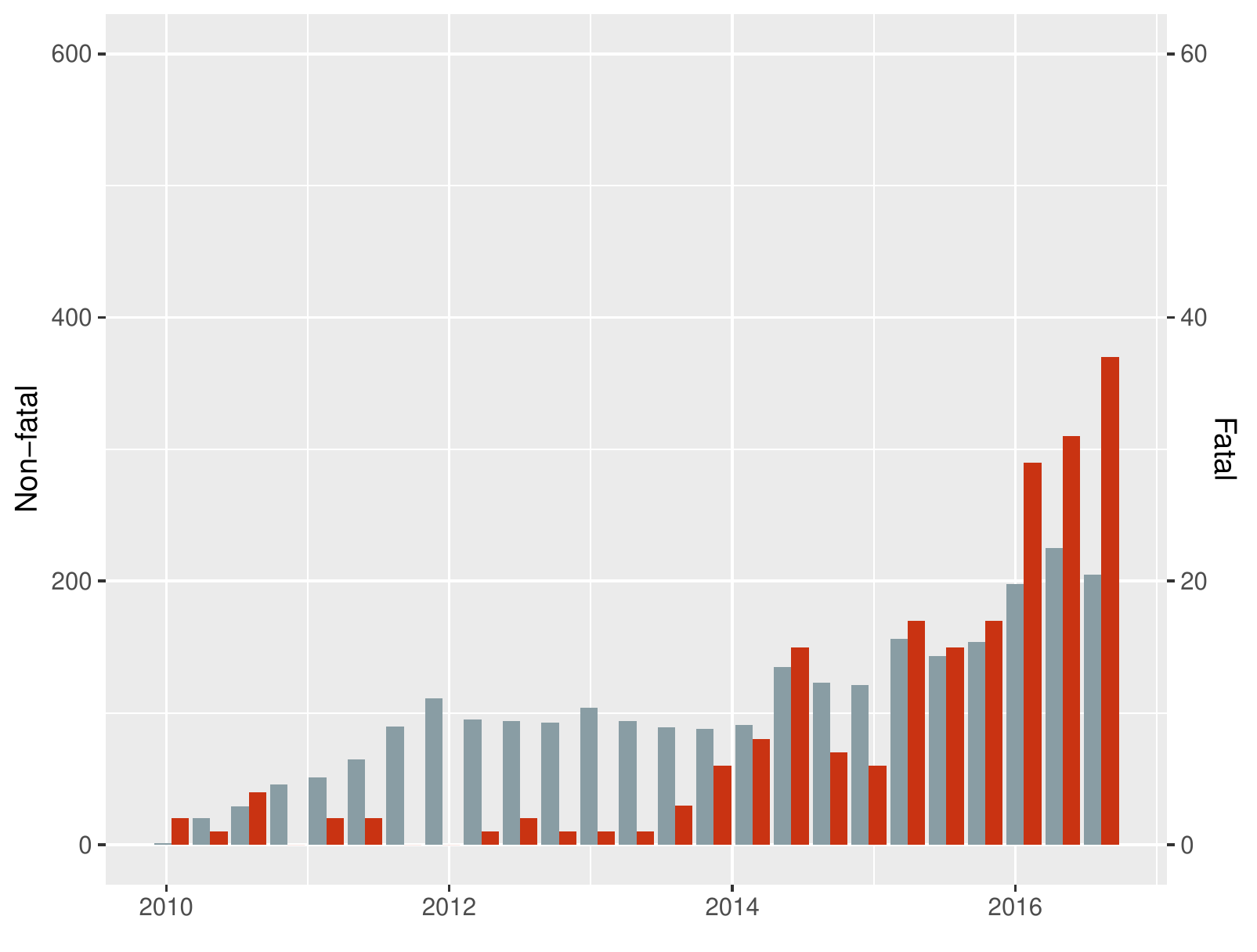}
	\label{fig:hist4}
	\end{subfigure}
	\caption{Histograms of non-fatal (grey) and fatal (red) overdose events for each group over time: group 1 (top left), group 2 (top right), group 3 (lower left), and group 4 (lower right)..}
	\label{fig:hist}
\end{figure}

 \begin{table}
\centering
 \begin{tabular}{c | l  l l l}
 \hline
 drug & group 1  & group 2 & group 3 & group 4 \\
 \hline
 1 & 6-MAM & Benzodiazepine&Ethanol  & Fentanyl  \\
 2 & Heroin& Hydrocodone &Alcohol & Norfentanyl\\
 3 &  Codeine&  Oxycodone & Cocaine & Opioids \\
 4 &  Morphine& Hydromorphone&  Illicits& Amphetamine \\
 5 & Illicit & Oxymorphone &  Benzodiazepine & Methamph. \\ 
 \hline
 \end{tabular}
 \caption{Top 5 drugs from each group.}
 \label{topictable}
 \end{table}
Next we fit the point process model to the fatal and non-fatal overdose data.  In Figure \ref{fig:hms} we plot a heatmap of the inferred background events in space, disaggregated by group, along with the temporal trend 
of background events in Figure \ref{fig:hist}.  We find that in time, the frequency of prescription opioid overdoses went down in Indianapolis, whereas illicit opioid overdoses, including the fentanyl group, increased over the same time period.  In space, the illict drug hotspots are focused downtown, whereas the prescription opioid hotspots are more spread out in the city. 

In Table  \ref{params} we display the estimated point process parameters.  We see that for each group self-excitation plays a large role, where the branching ratio ranges from .72 to .98.  In Table \ref{EMSmeasure} and \ref{Opmeasure} we compare the log-likelihood values of the combined heterogeneous point process to baseline models estimated only on EMS or overdose death data.  Here we find that including the EMS data improves the AIC values of the model for opioid overdose death, and the overdose death data improves the AIC of the model for EMS events.

To assess the model with a metric that better mirrors how interventions might work, we run the following experiment.  For each day in January 15, 2010 to December 30, 2016, we estimate the point process intensity in each of 50x50 grid cells covering Indianapolis.  We then rank the cells by the intensity and assign labels for whether an overdose occurs (1) or does not occur (0) during the next day.  We then compute the area under the curve (AUC) of this ranking for the baseline and the proposed method.  In practice, a point process model could be used to rank the top hotspots where overdoses are likely to occur and then those areas could be the focus of targeted interventions, such as distribution of naloxone that reverses the effects of an overdose.   

In Table \ref{EMSmeasure} and \ref{Opmeasure} we find that the AUC of the combined model evaluated on overdose death data is .85, compared to .81 for the model utilizing only overdose data.  However adding overdose death data to the EMS data impairs the model in terms of AUC. The heterogeneous model has an AUC of .72 compared to .8 for the EMS data model (though the overdose death data does improve the AIC of the EMS data model). 
 
 \begin{table}[h]
\centering
 \begin{tabular}{l | c| c | c |c}
 \hline
model & log-likelihood&  df & AIC & AUC\\
 \hline
baseline model&  $4.9892 \times 10^4$ &4 & $-9.9774 \times 10^4$ &{\bf 0.8032}\\
proposed model& $ 5.5752\times 10^4$ & 16 & $\mathbf{-1.1147\times 10^5}$& 0.7159\\
 \hline
 \end{tabular}
 \caption{Different measurement results on EMS data. }
 \label{EMSmeasure}
 \end{table}
 
\begin{table}[h]
\centering
 \begin{tabular}{l | c| c | c | c}
  \hline
model & log-likelihood & df & AIC & AUC\\
 \hline
baseline model& $-3.6110 \times 10^3$ & 16 & $7.2540 \times 10^3$ & 0.8088\\
proposed model & $1.7165 \times 10^3$ & 16 &  $\mathbf{-3.4009\times 10^3}$ &{\bf 0.8524}\\
\hline
 \end{tabular}
 \caption{Different measurement results on Opioid overdose death data. }
 \label{Opmeasure}
 \end{table}

 \begin{table}
\centering
 \begin{tabular}{c | l | l | l | l  }
 \hline
 Group  $\#$  & $K_0$  & w & $\mu$ & $\sigma$ \\
 \hline
 1 & 0.9609 & 0.0153 & 4.0517 & 0.0148 \\
 2 & 0.9864 & 0.0170 & 2.8304 & 0.0313 \\
 3 & 0.7257 & 0.0094 & 28.7279 & 0.0044 \\
 4 & 0.9214 & 0.0143 & 4.4550 & 0.0091 \\
 \hline
 \end{tabular}
 \caption{Parameters of estimated model for each group.}
 \label{params}
 \end{table}

%  It is also interesting that the log-likelihood trend is not monotonic as the percentage of $A$ goes bigger.
 
\section{Discussion}
% Fusing the labelled data with unlabelled data is useful in filling the labels for unlabelled data. Our experiment results have shown that the proposed point process model works better than a simpler model without clustering. 

% I expect the result to be worse, since the two datasets might not follow the same pattern, and also the unlabelled data is 20 times more than the labelled data. As shown in \cite{Cozman2003}, unlabeled data can lead to an increase in classification error even in situations where additional labeled data would decrease classification
% error. Although this is not showing in the simulation result, it might be the case on the real data. 

% Lastly, \cite{Zhuang2018} has shown that the periodical terms (e.g., hour, month, and year) can be added to the model to replace the trend term $t$. I would like to see if adding the periodical terms improve the performance of the model. If so, the missing hour information of drug abuse data can also be learned from EMS data.

Heterogeneous data integration for model improvement promotes several policy and intervention benefits. Research using emergency medical services data has shown that persons who experience repeat non-fatal drug overdoses have a significantly higher mortality rate as compared to individuals without repeat events \cite{ray2018ems}. As our results suggest, toxicology data can be leveraged to model overdose diffusion across space and time, and diffusion varies across geographies. Taken together, integration of large-scale event data and overdose diffusion can sharpen policy interventions designed to reduce substance abuse and substance-related deaths. One such policy example is the deployment of nasal naloxone by police and EMS agencies which mitigates overdose effects \cite{fisher2016police}. 

Integration of heterogeneous data sources also help to contextualize and better understand the nuances of how social harms may affect different populations of people. As our study illustrates, prescription drug overdoses occur at higher rates in areas further from downtown Indianapolis, while illicit drug overdoses are more concentrated around the urban core of the city. These results underscore societal differences of opioid drug use. Consistent with community explanations of crime and social disadvantage \cite{sampson1989community}, we observe that illicit drugs, which are more likely to result in mortality, may disproportionately impact minority communities.  Current evidence indicates these trends are driven by heroin and synthetic opioid-related deaths as well as growing use of fentanyl-laced cocaine among African Americans \cite{alexander2018trends,jalal2018changing}. Moreover, these trends persist despite evidence that African Americans are less likely to be prescribed opioids for pain relative to Caucasians \cite{meghani2012time}, which has been identified as a primary pathway to illicit opioid use \cite{mars2014every}. Together, current evidence suggests the epidemiology of opioid use, especially illicit opioid use, is not well-defined for racial-ethnic minorities. Heterogeneous data integration is likely the most appropriate path forward to improve our understanding of this issue.  

Our work here is also related to the analysis of free text data that accompanies crime reports \cite{kuang2017crime,pandey2018evaluation,mohler2018privacy} and other types of incidents, for example railway accidents
\cite{heidarysafa2018analysis}.  While the majority of point process focused studies of crime and social harm use only location, time, and incident category as input into the model, we believe future research efforts on incorporating auxilliary, high-dimensional information into these models may yield improvements in model accuracy and also provide insight into the underlying causal mechanisms in space-time event contagion.  

We do note that disentangling contagion patterns from other types of spatio-temporal clustering is challenging due to seasonal and exogeneous trends \cite{Zhuang2018,mohler2013modeling}.  Future work should also focus on investigating the extent to which drug overdose triggering found in the present study can be detected across cities and model specifications.

\section{Acknowledgements}

This work was supported in part by NSF \\
 grants ATD-1737996 and SCC-1737585. 

\bibliographystyle{imsart-nameyear}

\bibliography{main}

\end{document}